\documentclass[letterpaper, 10 pt, conference]{ieeeconf}  

\IEEEoverridecommandlockouts                              

\usepackage{graphicx} 
\usepackage{wrapfig}
\usepackage{amsmath}
\usepackage{amsfonts}
\usepackage{amssymb}
\usepackage{bbm}
\usepackage{xcolor}
\usepackage{url}
\usepackage{cite}
\usepackage{dsfont}
\usepackage{braket}
\usepackage{booktabs}
\usepackage{color}
\usepackage{colortbl}
\usepackage{mathtools}
\usepackage{multirow}
\usepackage{footmisc}
\usepackage{lmodern}

\title{\LARGE \bf
Tactile-Sensitive NewtonianVAE for High-Accuracy \\ Industrial Connector Insertion
}

\author{Ryo Okumura$^{*,\star}$, Nobuki Nishio$^{\dag}$ and Tadahiro Taniguchi$^{\dag,\ddag}$ 
    \thanks{$^{*}$ Ryo Okumura and Tadahiro Taniguchi are with Digital\&AI Technology Center, Technology Division, Panasonic Holdings Corporation, Japan.}%
    \thanks{$^{*}$ Nobuki Nishio is with R\&D Division, Panasonic Connect Co., Ltd., Japan.}%
    \thanks{$^{\ddag}$ Tadahiro Taniguchi is also with Ritsumeikan University, College of Information Science and Engineering, Japan.}%
    \thanks{$^{\star}$ Corresponding author: \texttt{okumura.ryo001@jp.panasonic.com}}
    }
 
\begin{document}

\maketitle
\thispagestyle{empty}
\pagestyle{empty}

\begin{abstract}
An industrial connector insertion task requires submillimeter positioning and grasp pose compensation for a plug.
Thus, highly accurate estimation of the relative pose between a plug and socket is fundamental for achieving the task. 
World models are promising technologies for visuomotor control because they obtain appropriate state representation to jointly optimize feature extraction and latent dynamics model.
Recent studies show that the NewtonianVAE, a type of the world model, acquires latent space equivalent to mapping from images to physical coordinates. 
Proportional control can be achieved in the latent space of NewtonianVAE.
However, applying NewtonianVAE to high-accuracy industrial tasks in physical environments is an open problem.
Moreover, the existing framework does not consider the grasp pose compensation in the obtained latent space.
In this work, we proposed tactile-sensitive NewtonianVAE and applied it to a USB connector insertion with grasp pose variation in the physical environments.
We adopted a GelSight-type tactile sensor and estimated the insertion position compensated by the grasp pose of the plug.
Our method trains the latent space in an end-to-end manner, and no additional engineering and annotation are required.
Simple proportional control is available in the obtained latent space.
Moreover, we showed that the original NewtonianVAE fails in some situations, and demonstrated that domain knowledge induction improves model accuracy.
This domain knowledge can be easily obtained using robot specification and grasp pose error measurement.
We demonstrated that our proposed method achieved a 100\% success rate and 0.3 mm positioning accuracy in the USB connector insertion task in the physical environment.
It outperformed SOTA CNN-based two-stage goal pose regression with grasp pose compensation using coordinate transformation.
\end{abstract}


\section{Introduction}

\begin{figure}[t!]
    \centering
    \vspace{0mm}
    \includegraphics[width=8.4cm]{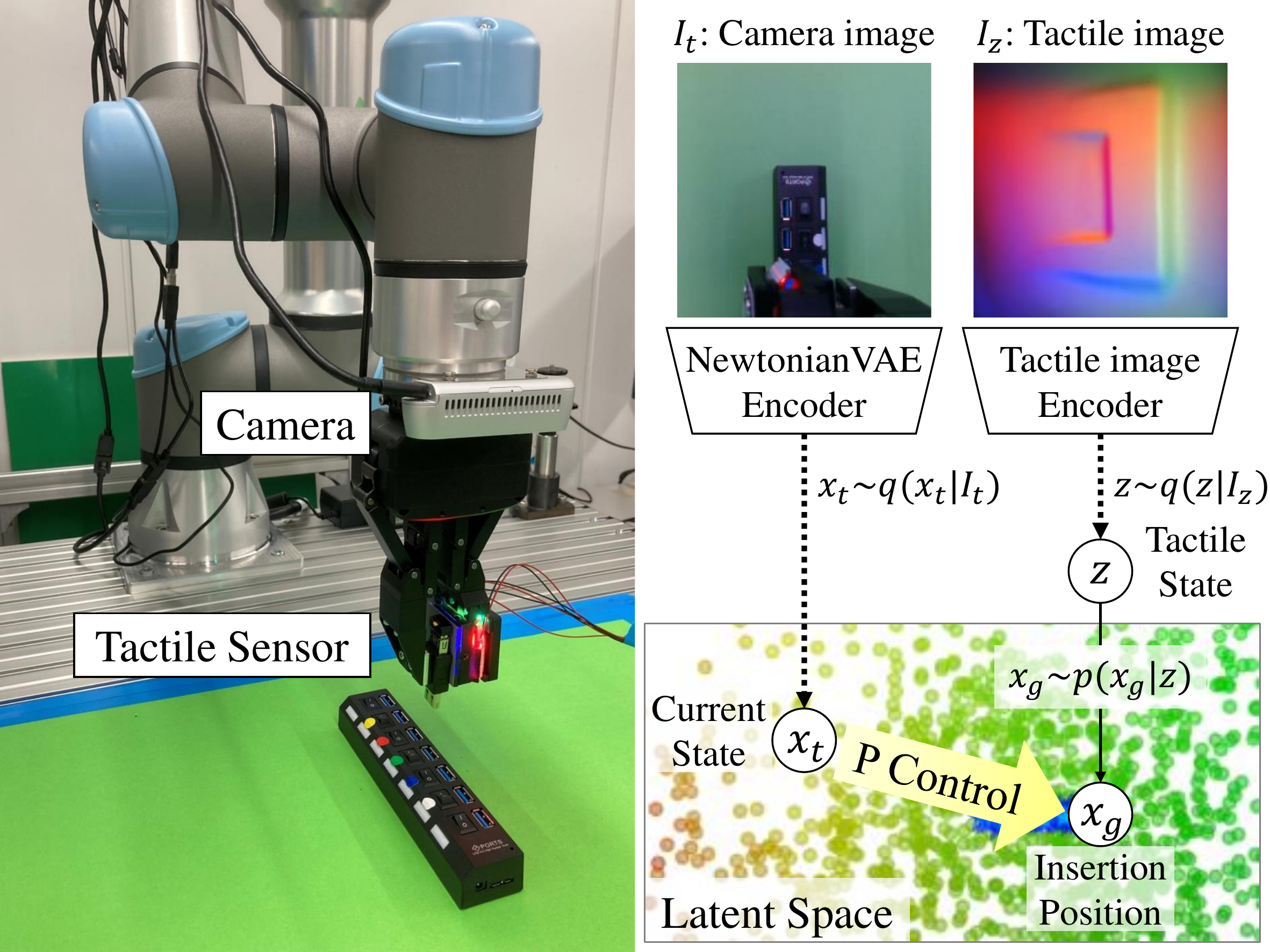}
    \vspace{-2mm}
    \caption{
    Overview of the proposed robot control system for connector insertion.
    An underactuated two-finger gripper, RGB camera, and GelSight-type tactile sensor are attached to an end of a robot arm.
    Latent states $\mathbf{x}_t$ and $\mathbf{z}$ are inferred from RGB camera and tactile sensor images, respectively.
    An insertion position $\mathbf{x}_g$ is generated from $\mathbf{z}$.
    A grasped plug is positioned to the insertion position by proportional control in the latent space, even with grasp pose variations.
    }
    \label{fig:overview}
    \vspace{-2mm}
\end{figure}

Positioning is one of the fundamental skills in factories, warehouses, retail stores, etc.
It is required in industrial operations like insertion, fitting, pick and place, and so on.
Generally, poses of target objects like holes, sockets, dowel pins, and so on, are unknown.
Thus, target poses must be localized from observations like camera images.
Recent studies show that robot learning can automate these positioning tasks in a data-driven manner \cite{ResidualRL, SSL4P3}.

The robot learning technology with image recognition is promising because hand-designed image feature extraction requires high levels of expertise.
Using deep learning-based image recognition techniques, target object positions can be estimated from camera images without feature engineering \cite{Johns2021CoarsetoFineIL}.

World models \cite{ha:worldmodels} are also a promising technology for visuomotor control since appropriate state representation for control is obtained by optimizing feature extraction and latent dynamics models.
The NewtonianVAE \cite{NVAE}, a kind of the world models, acquires latent space that is equivalent to mapping image observations to physical coordinates.
The obtained state representation is appropriate for the positioning tasks. 
In many positioning tasks, compensation for grasp pose variation is a significant problem to be solved.
However, the NewtonianVAE does not have the mechanism to compensate for the grasp pose variation.

Precise estimation of the grasp pose is necessary for solving the above problem.
Nevertheless, estimating the grasp pose from the camera images is difficult because the grasped objects are occluded by grippers.
Tactile sensors enable pose estimation of the grasped objects.
Especially, GelSight-type tactile sensors \cite{GelSight} achieve the grasp pose estimation in submillimeter accuracy \cite{li2014localization}.

In this paper, we propose a method that maps end-effector-attached camera images onto latent space to position grasped objects, especially USB plugs.
The latent space is acquired via the NewtonianVAE, to have the same axes and scale as the physical coordinate system.
To compensate for the grasp pose variation, we adopted the GelSight-type tactile sensor, whose raw outputs are directly mapped to insertion position in the latent space, in which the plug is right above a socket.
We fed back the difference between current and insertion positions in the latent space and applied simple proportional control to position the plug above the socket.
The mapping from the camera and GelSight images onto the latent space was trained in an end-to-end manner.

Our contributions are in three-folds.
\begin{itemize}
    \item
    We proposed a general framework for positioning grasped objects to the goal position by visual feedback control using a tactile-sensitive NewtonianVAE (TS-NVAE).
    We adopted the GelSight-type tactile sensor to compensate for the grasp pose variation in the latent space.
    TS-NVAE requires no additional feature engineering, annotation, or calibration.
    \item
    We demonstrated that the original NewtonianVAE fails in some situations, and domain knowledge about transition uncertainty and grasp pose variation improves model accuracy.
    Furthermore, this domain knowledge is easy to know from robot specification and grasp pose error measurement.
    \item
    We demonstrated that TS-NVAE applies to industrial insertion tasks requiring submillimeter positioning accuracy in physical robot control systems.
    It achieved 0.3 mm accuracy and outperformed SOTA CNN-based two-stage goal pose regression with grasp pose compensation using coordinate transformation.
\end{itemize}

TS-NVAE has several properties desirable for the industrial application in physical environments.
First, it achieves 0.3 mm positioning accuracy which is enough for many industrial tasks like assembly, palletizing, mounting, etc.
Second, we can optimize the trade-off between accuracy and tact time by tuning the proportional control parameters.
Third, the training is sample efficient, fast, and fully offline.
We needed only 300 s of transition data for training and 13 hours of training time in our experiments.
Finally, TS-NVAE is explainable, because the latent space have the same axes and scale as the physical space.

\section{Related Work}

\subsection{World Models}

World model \cite{ha:worldmodels} is a generic term for models including state inference and transition in partially observable Markov decision processes (POMDPs).
Joint optimization of the inference and transition models leads to accurate and sample-efficient training.
There are three types of control methodology using the trained world models, policy, MPC, and optimal control.

The policy-based method jointly or separately trains policy besides world models.
Lee et al. \cite{lee2019slac} trained the policy using the actor-critic method according to the control as inference theory \cite{CaI}.
Hafner et al., Okada et al. and Kinose et. al. \cite{dreamer, dreaming, dreamingv2, mvdreaming} trained the policy from virtually collected transition data in world models using the overshooting technique \cite{hafner2018planet}.
These methods require reward definitions because the policy is optimized to maximize cumulative rewards in an episode. 
In the connector insertion tasks, measuring the relative pose between the plugs and sockets is difficult, thus dense rewards are generally unavailable.
However, sparse rewards lead to low sample efficiency.
Additionally, these methods do not work in offline learning settings due to the difference between learned and behavior policies \cite{CQN}.
Rafael et al. \cite{LOMPO} alleviated this problem to quantify epistemic uncertainty and induced an uncertainty penalty term into the rewards. 

Hafner et al., Okada et al. and Okumura et. al. \cite{hafner2018planet, planetofbayesians, dacssm} used MPC \cite{Garcia:1989:MPC:72068.72069} in the obtained latent space and evaluated performance in MuJoCo \cite{Todorov2012MuJoCoAP} simulator environments.
MPC infers optimal action sequence by predicting future cumulative rewards, given current states.
Since it also needs access to rewards, the same problem as the policy-based method occurs.
Moreover, MPC tends to fail with the sparse rewards because the future prediction gets no rewards within prediction horizons.
However, extending the prediction horizons needs high computational costs, leading to low control frequencies.

Levine et al. and Watter et al. \cite{gps, ContactRich, e2c} applied optimal control in the latent space.
They induced locally linear constraints into the transition models of the world models.
They used off-the-shelf stochastic optimal control methodology like iterative linear quadratic regulator (iLQR) to minimize the cost function.
Note that the cost function must follow a fixed formula, e.g. quadratic function for iLQR.
For positioning tasks, square distances to the goal are suitable for the quadratic cost function, but generally, we do not have access to the distance in the physical environments.

These existing works above need access to rewards which lead to restricted application in the physical environments.
Especially, distance-based dense rewards are desirable for positioning tasks, but difficult to obtain.
Therefore, the goal conditional control method is preferable for the positioning tasks.
Yan et al. \cite{wilsoncontrastive} used goal conditional MPC and achieved manipulation of deformable objects.
They employed similarity function, distance in the latent space between current and goal states, for objectives of MPC. 
Angelina et al. \cite{visualplanning} adopted the inverse dynamics model to infer action from a pair of sequential observations.
They applied constraints by contrastive predictive coding \cite{cpc} or causal InfoGan \cite{wang:infogan} to map sequential observation pairs to a neighborhood in the latent space.
Our proposed method employed the proportional control whose goal state is conditioned on the tactile information.

\subsection{Tactile Sensors}

Tactile sensors are important devices in positioning tasks to precisely estimate the in-hand grasp pose. 
In this section, we introduce tactile sensors that have soft sensing surfaces for contact state perception.
Wettels et al. \cite{biotac} developed biomimetic tactile sensors having multimodal sensor arrays.
Lammers et al. \cite{TakkTile} used MEMS microphones to sense contact forces and paved them under the sensing surface. 
In the works above, the distance between each sensor element is several millimeters.
They do not have sufficient accuracy for submillimeter positioning tasks.

Optical tactile sensors \cite{kamiyama2004gelforce} use cameras to take images of the sensing surface deformation.
Especially, GelSight \cite{GelSight} achieved sufficient accuracy for the grasp pose estimation for connector insertion \cite{li2014localization}, thanks to the high-resolution sensor images.
They used conventional keypoints matching to localize the in-sensor contact pose of the plug, assuming no displacement out of the sensing surface of the tactile sensor.
Then, they transformed the in-sensor pose into a Cartesian coordinate.
For accurate coordinate conversion, the relative pose between the robot and tactile sensor must be known. 
Thus, it was measured using a calibration jig. 
However, it is problematic to use the sensor on the gripper finger, especially for an underactuated gripper, because the relative pose between the robot and sensor changes depending on the opening width of the gripper’s fingers. 
Unlike the existing work above, our proposed method needs no jig. 
We need not assume that grasped objects have no displacement out of the sensor surface.

\section{Preliminaries}

\subsection{NewtonianVAE}

NewtonianVAE is an instance of the world model. 
The most important property of the NewtonianVAE is that proportional control is possible in the latent space.
The NewtonianVAE assumes POMDPs settings where $\mathbf{I}_t$ is observation image, $\mathbf{x}_t \in \mathbb{R}^D$ is the latent state, and $\mathbf{u}_t \in \mathbb{R}^D$ is action.
Note that $\mathbf{x}_t$ and $\mathbf{u}_t$ have the same degrees of freedom $D$ to satisfy a formulation discussed below.

They modeled the latent transition with double integrator dynamics, inspired by Newton's equations of motion:
\begin{eqnarray}
   \frac{d\mathbf{x}}{dt} &=& \mathbf{v} \nonumber \\
   \frac{d\mathbf{v}}{dt} &=& A(\mathbf{x}, \mathbf{v}) \cdot \mathbf{x} + B(\mathbf{x}, \mathbf{v})\cdot \mathbf{v} + C(\mathbf{x}, \mathbf{v})\cdot \mathbf{u} \nonumber
   \label{eq:act_motion}
\end{eqnarray}

They trained the world models to maximize the variational lower bound below:
\begin{eqnarray}
\mathbb{E}_{q(\mathbf{x}_t|\mathbf{I}_t)q(\mathbf{x}_{t-1}|\mathbf{I}_{t-1})} [ \mathbb{E}_{p(\mathbf{x}_{t+1}|\mathbf{x}_t, \mathbf{u}_t; \mathbf{v}_{t+1})} \log p(\mathbf{I}_{t+1}|\mathbf{x}_{t+1}) \nonumber \\
- \mathrm{KL}\left(q(\mathbf{x}_{t+1}|\mathbf{I}_{t+1}) \|  p(\mathbf{x}_{t+1}|\mathbf{x}_t, \mathbf{u}_t; \mathbf{v}_{t+1}) \right) ]
\label{eq:nvae_elbo}
\end{eqnarray}
where the transition prior is:
\begin{eqnarray}
p(\mathbf{x}_{t+1}|\mathbf{x}_t, \mathbf{u}_t; \mathbf{v}_{t+1}) = \mathcal{N}(\mathbf{x}_{t+1}|\mathbf{x}_t + \Delta t \cdot \mathbf{v}_{t+1}, \; \sigma^2) \label{eq:nvae_trans} \\
\mathbf{v}_{t+1} = \mathbf{v}_t + \Delta t \cdot  (A \mathbf{x}_t + B \mathbf{v}_t + C \mathbf{u}_t) \label{eq:nvae_vt}
\end{eqnarray}
with
\begin{eqnarray}
A &=& \mathrm{diag}(f_A(\mathbf{x}_t, \mathbf{v}_t, \mathbf{u}_t)) \nonumber \\
\log(-B) &=& \mathrm{diag}(f_B(\mathbf{x}_t, \mathbf{v}_t, \mathbf{u}_t)) \nonumber \\
\log C &=& \mathrm{diag}(f_C(\mathbf{x}_t, \mathbf{v}_t, \mathbf{u}_t)) \nonumber
\end{eqnarray}
where $f_A,\ f_B,\ f_C$ are neural networks with linear output activation.
$\mathbf{v}_t$ is computed as $\mathbf{v}_t = (\mathbf{x}_t - \mathbf{x}_{t-1})/\Delta t$.
Because transition matrices $A,\ B,\ C$ are diagonal, linear combinations between each latent dimension are eliminated.
Therefore, correct coordinate relations between $\mathbf{x}$, $\mathbf{v}$, and $\mathbf{u}$ are obtained.
For a physical robot experiment, they set $A=0$, $B=0$, and $C=1$ to improve sample efficiency.
An additional regularization term was placed on the latent space, $\mathrm{KL}(q(\mathbf{x}|\mathbf{I}) \| \mathcal{N}(0,1))$.

The proportional control like mixture density network \cite{bishop1994mixture} and dynamic movement primitives \cite{ijspeert2013dynamical, schaal2006dynamic} have been performed in the acquired latent space.

\subsection{GelSight}

GelSight is an optical tactile sensor composed of a camera, RGB colors of LEDs, transparent gel whose one side surface is covered with an opaque thin coating.
The LEDs’ light is emitted to the coating through the gel.
The camera captures images of the coating illuminated by the LEDs through the gel.
If objects are in contact with the other side of the coating surface, the gel and coating deform. 
Thus, the camera captures the contact surface deformation.
The obtained images are used for 3D reconstruction \cite{GelSight}.
Tian et al. used RGB images directly to control robots \cite{Tian2019ManipulationBF}.
In this paper, we employed raw RGB images of the GelSight-type tactile sensor to estimate the insertion pose.

\section{Tactile-Sensitive NewtonianVAE}

\subsection{Training Objective}

Our work is based on the NewtonianVAE. 
We jointly trained a goal state prediction model using tactile sensor images besides the NewtonianVAE latent model, given:
\begin{itemize}
\item image-action sequences $D_x = {(\mathbf{I}_1, \mathbf{u}_1), ..., (\mathbf{I}_T, \mathbf{u}_T )}$
\item a pair of images $D_z = {(\mathbf{I}_z, \mathbf{I}_g)}$
\end{itemize}
where $\mathbf{I}_t$ is a camera image, $\mathbf{I}_z$ is a tactile sensor image and $\mathbf{I}_g$ is a camera image at the insertion position.
The loss function of TS-NVAE is $\mathcal{L}_x+\mathcal{L}_z$ such that:
\begin{eqnarray}
\mathcal{L}_x &=& \mathbb{E}_{q(\mathbf{x}_t|\mathbf{I}_t)} \mathbb{E}_{p(\mathbf{x}_{t+1}|\mathbf{x}_t, \mathbf{u}_t)} [-\log p(\mathbf{I}_{t+1}|\mathbf{x}_{t+1}) \nonumber  \\
&& + \mathrm{KL}\left(q(\mathbf{x}_{t+1}|\mathbf{I}_{t+1}) \|  p(\mathbf{x}_{t+1}|\mathbf{x}_t, \mathbf{u}_t) \right) ]
\label{eq:Lx}
\end{eqnarray}
\begin{eqnarray}
\mathcal{L}_z &=& \mathbb{E}_{q(\mathbf{z}|\mathbf{I}_z)} \mathbb{E}_{p(\mathbf{x}_g|\mathbf{z})}[-\log p(\mathbf{I}_z|\mathbf{z}) \nonumber  \\
&& -\log p(\mathbf{I}_g|\mathbf{x}_g) + \mathrm{KL} \left(q(\mathbf{x}_{g}|\mathbf{I}_{g}) \|  p(\mathbf{x}_g|\mathbf{z}) \right)]
\label{eq:Lz}
\end{eqnarray}

We employed the built-in Cartesian velocity control function of the robot arm, i.e. $\mathbf{u}_t$ is a reference velocity in the Cartesian space.
We simplified the transition model of the NewtonianVAE as below:
\begin{eqnarray}
p(\mathbf{x}_{t+1}|\mathbf{x}_t, \mathbf{u}_t) = \mathcal{N}(\mathbf{x}_{t+1}|\mathbf{x}_t + \Delta t \cdot \mathbf{u}_t, \; \sigma_{x}^2)
\label{eq:transition_prior}
\end{eqnarray}
Due to the simplification, the latent space has the same axes and scales as the Cartesian coordinate system.

\subsection{Induce domain knowledge}
Assuming the latent and physical space have the same axes and scales, we can easily induce domain knowledge into the latent distribution.
First, $\sigma_{x}$ in the transition model represents the transition uncertainty of the robots.
We set $\sigma_{x}=0.0001$, that is nominal repeated positioning accuracy of the UR5e (0.1 mm). 

Moreover, we added the additional KL regularization terms to the training objective.
\begin{eqnarray}
\mathbb{E}_{q(\mathbf{z}|\mathbf{I}_z)} [ \mathrm{KL} \left(q(\mathbf{x}_{g}|\mathbf{I}_{g}) \| \mathcal{N}(0, \; \sigma_{g}^2) \right) \nonumber \\
+ \mathrm{KL} \left(p(\mathbf{x}_g|\mathbf{z}) \|  \mathcal{N}(0, \; \sigma_{g}^2) \right)]
\label{eq:additional_kl}
\end{eqnarray}
$\sigma_{g}$ represents the variation of the insertion position.
In the context of our problem settings, it is equivalent to the grasp pose variation.
In our experiments described below, $\sigma_{g}$ was set to 0.0015 so that the grasp pose variation ($\pm$3 mm) was equivalent to $\pm2\sigma_{g}$.

\section{Experiment}

\subsection{Hardware Configuration}

Fig. \ref{fig:overview} shows our experimental setup.
The robot arm is a 6-DoF Universal Robots UR5e.
The end-effector is a ROBOTIS RH-P12-RN(A) underactuated two-finger gripper. 
The Intel RealSense D435i RGBD camera was attached to a tool mount of the UR5e.
In our experiment, depth images were not used.
We attached a GelSight-type tactile sensor to one finger of the gripper, and a target USB hub with seven ports to a lab bench.
We used only the center port of the USB hub.
Also, we used only the camera and tactile images for control, so the controller had no access to the Cartesian nor joint pose of the robot arm.
In all experiments, we used the built-in Cartesian velocity control in the X and Y axes in the UR5e robot arm.

\subsection{Data collection}

\begin{figure}[t!]
    \centering
    \vspace{1mm}
    \includegraphics[width=8.4cm]{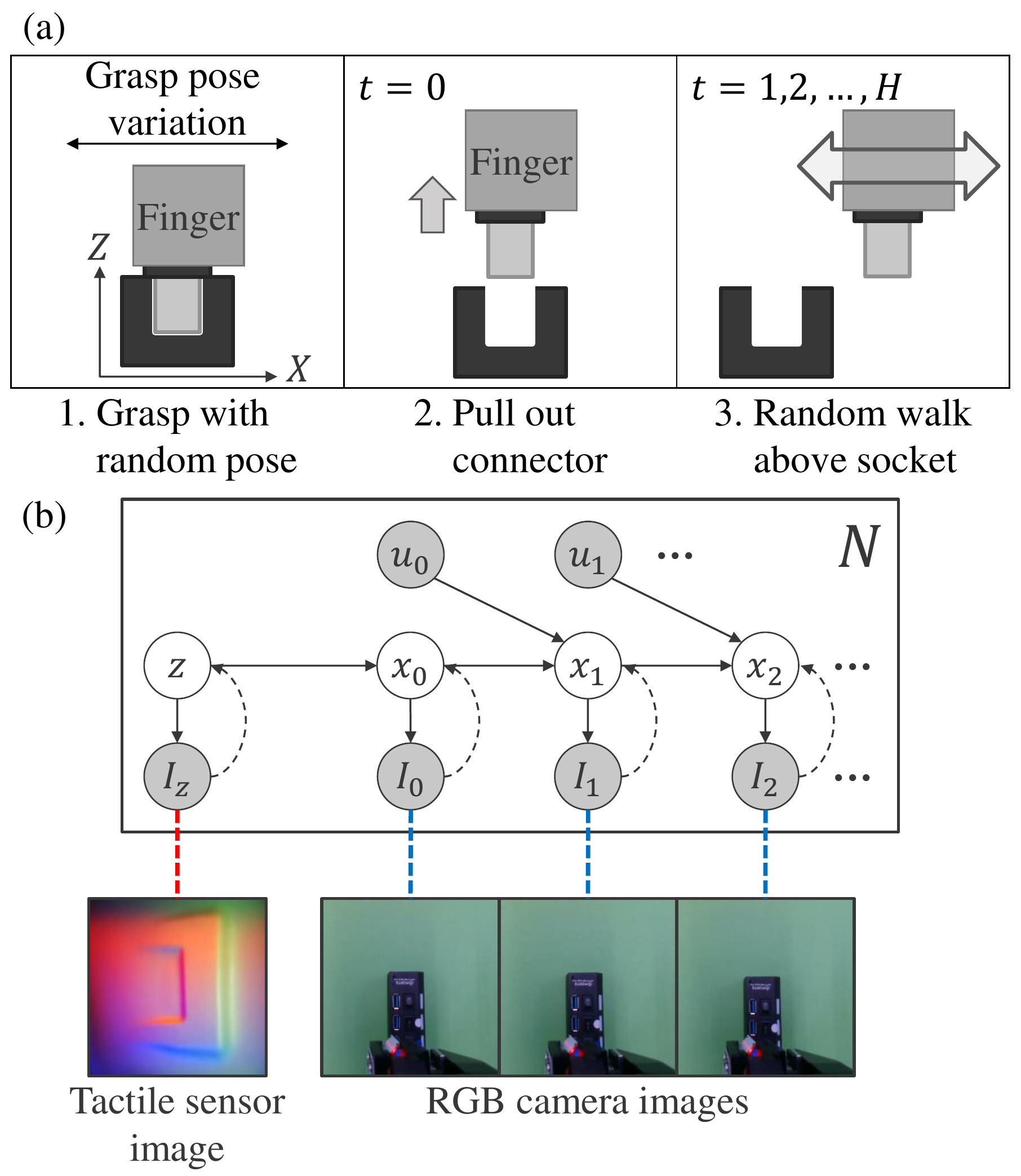}
    \vspace{-3mm}
    \caption{(a) Data collection process. First, a robot grasps a plug with a random pose. Second, it pulls out the plug upward. In this time, we observe $\mathbf{I}_z$ and $\mathbf{I}_0$. $\mathbf{I}_0$ is treated as $\mathbf{I}_g$ during training of insertion position estimator $p(\mathbf{x}_g|\mathbf{z})$. Then the robot randomly walks above a socket to collect transition data. Time horizon $H$ is 20, episode number $N$ is 30. (b) Graphical model. $\mathbf{x}_0$ is a latent state at the insertion position, and generated from $\mathbf{z}$.}
    \label{fig:data_collection}
    \vspace{-3mm}
\end{figure}

Fig. \ref{fig:data_collection} (a) shows our data collection process.
The robot pulls out the USB plug from the socket and observes the first camera and tactile images $\mathbf{I}_{0}$ and $\mathbf{I}_{z}$, respectively. 
At this time, the robot should be at the insertion position corresponding to the grasp pose variation that can be observed by $\mathbf{I}_{z}$.
Thus, $\mathbf{I}_{0}$ is treated as $\mathbf{I}_{g}$ during training of the insertion position estimator $p(\mathbf{x}_g|\mathbf{z})$.
Then, the robot randomly walks above the USB hub and collects transition data tuple $(\mathbf{I}_{t}, \mathbf{u}_{t})$.
The random action was uniformly sampled in the range of $\pm$0.01 m/s in the X and Y direction.
This data collection process assures that the insertion position $\mathbf{x}_{g}$ exists in the beginning of the episodes, leading to training positional relationship between the insertion position $\mathbf{x}_{g}$ and other position $\mathbf{x}_{t}$ via the transition model Eq. (\ref{eq:transition_prior}).
We collected 30 trajectories for the experiments.
One episode had 10 s length, and a control frequency was 2 Hz, i.e. one episode had 20 transitions.
We collected only 300 s and 600 transitions for training.

\subsection{Graphical model}
Fig. \ref{fig:data_collection} (b) shows the proposed graphical model whose observations are the camera images $\mathbf{I}_t$ and tactile image $\mathbf{I}_z$.
The latent states are $\mathbf{x}_t$ and $\mathbf{z}$, which generate $\mathbf{I}_t$ and $\mathbf{I}_z$, respectively.
As discussed in the previous section, $\mathbf{x}_0$ is always the insertion position, $\mathbf{x}_g$.
Thus, $\mathbf{x}_0$ is generated from the tactile latent state, $\mathbf{z}$.

\subsection{Training}
We used $224\times224$ and $64\times64$ pixels RGB images as inputs of the camera and tactile encoders, respectively.
ResNet \cite{resnet} with 18 layers was employed as the camera image encoder and pretrained on ImageNet \cite{imagenet}.
We used a simple CNN as the tactile image encoder.
Note that we did not employ a pretrained model for the tactile encoder because the tactile images have large domain shift in the ImageNet and other large-scale image datasets.

The output of the last layer of the camera and tactile encoder was fed into a fully connected layer to infer the posterior.
We adopted fully connected networks with two hidden layers of 16 units and the LeakyReLU activation to infer $\mathbf{x}_t$ and $\mathbf{z}$ from the encoded features, and the insertion position $\mathbf{x}_g$ from $\mathbf{z}$.
All posterior and prior were modeled as Gaussian distribution. 
We inputted eight full sequences as a mini-batch and executed 200 k steps of training (About 50 k epochs).
All models were trained using Adam \cite{adam} with a learning rate of $3\times10^{-4}$.

\begin{figure*}[t!]
    \centering
    \vspace{3mm}
    \includegraphics[width=17.7cm]{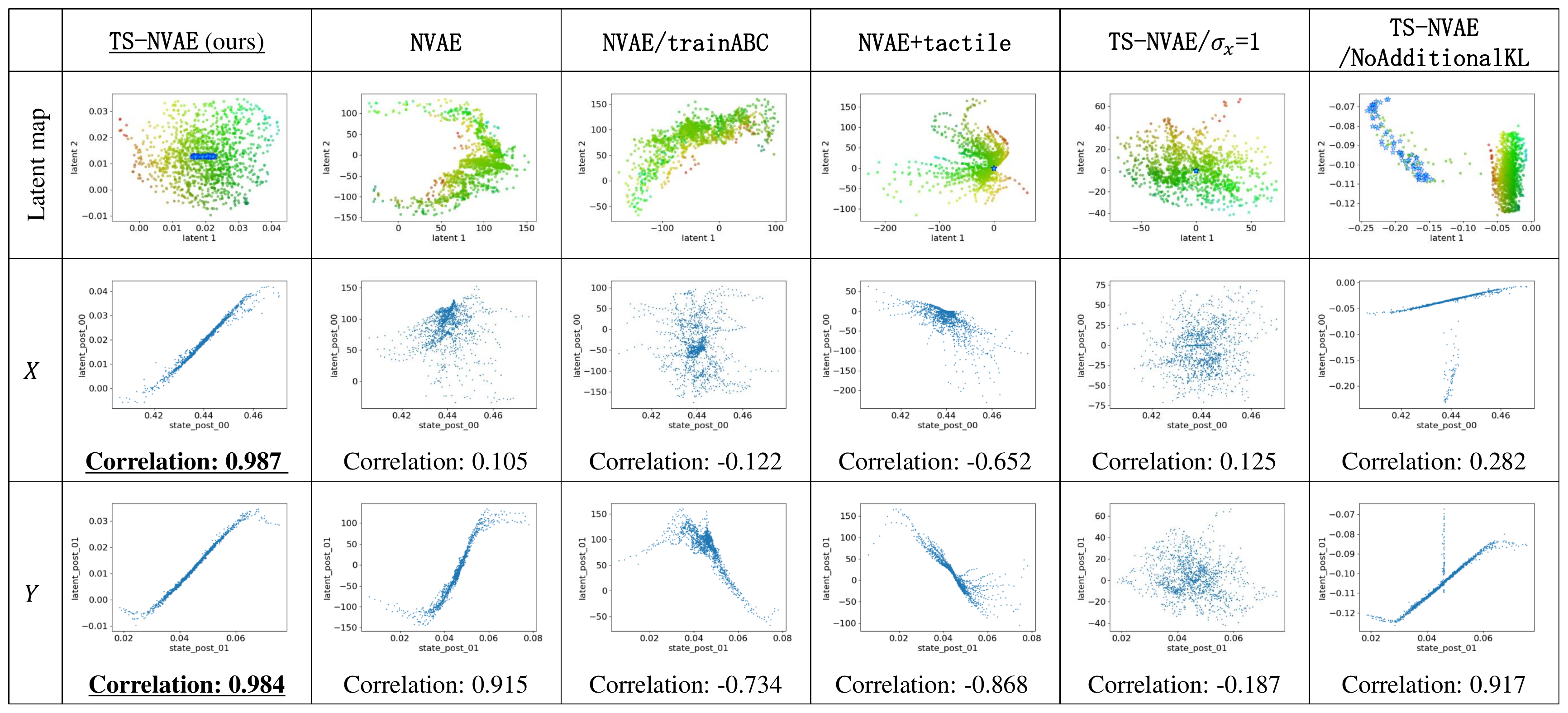}
    \caption{
    Visualization of latent spaces.
    The top row shows latent maps, where the horizontal and vertical axes are the first and second dimensions of the latent states, respectively.
    The marker color represents the position in the Cartesian coordinate system.
    Color variation from red to green represents the $X$ position in the Cartesian coordinate system.
    Color saturation represents $Y$, where high saturation indicates a high $Y$ position.
    Blue stars represent the estimated insertion positions from the tactile images.
    For \texttt{NVAE} and \texttt{NVAE/trainABC}, insertion positions are not plotted because tactile sensor-related models were untrained.
    The middle and bottom rows show the correlation between the $X$ and $Y$ positions in the Cartesian coordinate and the corresponding latent states, respectively.
    \texttt{TS-NVAE} is the proposed method.
    \texttt{NVAE} represents naive NewtonianVAE whose transition matrices are fixed to $A=0$, $B=0$, and $C=1$.
    In \texttt{NVAE/trainABC}, the transition matrices are trainable.
    In \texttt{NVAE+tactile}, a training objective includes $\mathcal{L}_z$.
    \texttt{TS-NVAE/$\mathbf{\sigma}_x$=1}, and \texttt{TS-NVAE/NoAdditionalKL} represents the ablation studies.
    In \texttt{TS-NVAE/$\mathbf{\sigma}_x$=1}, the domain knowledge on transition uncertainty was not used and we set $\mathbf{\sigma}_x=1$.
    In \texttt{TS-NVAE/NoAdditionalKL}, we did not use the additional KL regularization term in Eq. (\ref{eq:additional_kl}) for training.
    }
    \label{fig:latent_visualization}
    \vspace{-1mm}
\end{figure*}

\subsection{Experiment 1: Visualization of latent space}

After training, we visualized the acquired latent space and compared our proposed method with some instances of the NewtonianVAE baselines and ablation studies (Fig. \ref{fig:latent_visualization}).
The top row of Fig. \ref{fig:latent_visualization} shows latent maps. 
Variation of marker color hue and saturation represent the $X$ and $Y$ positions of the Cartesian coordinate, respectively. 
The blue star represents estimated insertion positions from the tactile images. 
The middle and bottom rows show the correlation between the $X$ and $Y$ position and corresponding latent state, respectively.
In all experiments, 70 validation episodes were collected using the procedure that was used for acquiring the training data.

Here, we evaluated three types of NewtonianVAE baselines.
\texttt{NVAE} is the naive implementation of NewtonianVAE with transition matrices fixed to $A=0$, $B=0$, $C=1$.
In \texttt{NVAE/trainABC}, the transition matrices $A,\ B$ and $C$ are trainable.
Note that the tactile sensor information was not used for these two baselines; thus, the grasp pose variation was not considered.
In \texttt{NVAE+tactile}, tactile images were employed for the compensation, and the training objective included $\mathcal{L}_z$ in Eq. (\ref{eq:Lz}).

For all NewtonianVAE baselines, the action $\mathbf{u}_t$ was acceleration. 
Here, we employed the original NewtonianVAE training objective (Eqs. (\ref{eq:nvae_elbo}), (\ref{eq:nvae_trans}), and (\ref{eq:nvae_vt})).
To control the UR5e robot arm, we calculated the reference velocity, $\mathbf{v}_t$, from the acceleration action $\mathbf{u}_t$ based on Eq. (\ref{eq:nvae_vt}).
No domain knowledge was induced into the NewtonianVAE instances.

\texttt{TS-NVAE/$\mathbf{\sigma}_x$=1} and \texttt{TS-NVAE/NoAdditionalKL} are ablation studies.
In the ablation studies, the action $\mathbf{u}_t$ was velocity, and the simplified transition model (Eq. (\ref{eq:transition_prior})) was used. 
In \texttt{TS-NVAE/$\mathbf{\sigma}_x$=1}, we did not use the domain knowledge on the transition uncertainty, i.e., $\mathbf{\sigma}_x$ was set to 1.
In \texttt{TS-NVAE/NoAdditionalKL}, we did not use the domain knowledge on the grasp pose variation, i.e., the additional KL regularization terms (Eq. (\ref{eq:additional_kl})) were not used in the training process.

The latent state of the proposed \texttt{TS-NVAE} method demonstrated a high correlation with the physical coordinate system. 
The estimated insertion positions are located near the center of the latent space of \texttt{TS-NVAE}.
This is considered reasonable because we collected data using a random walk process beginning from the insertion position.
In the three NewtonianVAE baselines and \texttt{TS-NVAE/$\mathbf{\sigma}_x$=1}, the correlations between the physical and latent positions were low because the domain knowledge on transition uncertainty was not considered;
the transition uncertainty $\mathbf{\sigma}_x$ was set to 1, which is a much larger value than the positioning accuracy of the UR5e.
For \texttt{TS-NVAE/NoAdditionalKL}, the correlation was high (except at the insertion position); however, the insertion positions were not mapped correctly onto the latent space without the additional KL regularization terms.

\subsection{Experiment 2: Positioning performance}

We evaluated positioning accuracy using simple proportional control in the latent space.
Here, the reference velocity $\mathbf{u}_t$ to control the UR5e robot arm was calculated in a proportional control manner as follows:
\begin{eqnarray}
\mathbf{u}_t=\alpha(\mathbf{x}_g - \mathbf{x}_t) \nonumber
\label{eq:proportinal_control}
\end{eqnarray}
In this calculation, $\alpha$ was set to 1, and the control frequency was 20 Hz.
Table \ref{table:positioning_accuracy} shows the success rate of the USB connector insertion and positioning errors (mean and standard deviation).
The positioning error was distance between the final position of the robot and the insertion position in which the plug was just above the socket.

In addition to the evaluations of the baselines and the ablation studies discussed in the previous section, we also evaluated coarse-to-fine imitation learning (CFIL) \cite{Johns2021CoarsetoFineIL} and its corresponding instances.
CFIL is an SOTA positioning technology that employs two-stage (coarse and fine stages) CNN-based goal pose regression. 
Here, we implemented the grasp pose compensation using coordinate transformation \cite{li2014localization} onto the CFIL.
The insertion position for the gripper with respect to the robot's base frame ${}^bT_g$ was calculated as follows:
\begin{eqnarray}
{}^bT_g = {}^bT_o ({}^mT_o)^{-1} ({}^gT_m)^{-1}
\label{eq:coordinate_transformation}
\end{eqnarray}
where $T$ denotes a coordinate transformation like a homogeneous transformation matrix.
The subscripts and superscripts of $T$ represent target and reference frame, respectively.
${}^gT_m$ denotes the pose of the sensor origin relative to the gripper, which is measured from the CAD data.
${}^bT_o$ is the target plug pose in the robot base coordinate, calculated as follows:
\begin{eqnarray}
{}^bT_o = {}^bT_g^{\rm{CF}} \, {}^gT_m \, {}^mT_o^{\rm{EX}}
\label{eq:calc_bTo}
\end{eqnarray}
where ${}^bT_g^{\rm{CF}}$ is the insertion position \textbf{\underline{without}} grasp pose compensation, which is estimated by the CFIL,
and ${}^mT_o^{\rm{EX}}$ is the plug pose relative to the tactile sensor origin for an expert data in CFIL.
${}^mT_o$ is the current plug pose relative to the tactile sensor origin.
In \texttt{CFIL+TactileCNN}, ${}^mT_o$ was estimated using CNN-based regression.
Here, the regression model was trained in a supervised manner.
In \texttt{CFIL+Template}, ${}^mT_o$ was calculated using template matching.
Note that grasp pose compensation was not considered in naive \texttt{CFIL}.

\begin{table}[ht]
\vspace{3mm}
\caption{Connector insertion task performance \\ (accuracy is represented as mean $\pm$ standard deviation)}
\label{table:positioning_accuracy}
\begin{center}
\vspace{-5mm}
\begin{tabular}{lcc}
\toprule
Method & Success rate & Accuracy [mm] \\
\midrule
\texttt{TS-NVAE} (ours) & \textbf{100\%(40/40)} & \textbf{0.3$\pm$0.1} \\
\texttt{NVAE} & 0\% (0/10) & $\gg50$ \\
\texttt{NVAE/trainABC} & 0\% (0/10) & $\gg50$ \\
\texttt{NVAE+tactile} & 0\% (0/10) & $\gg50$ \\
\texttt{TS-NVAE/NoAdditionalKL} & 0\% (0/10) & $\gg50$ \\
\texttt{TS-NVAE/$\mathbf{\sigma}_x$=1} & 0\% (0/10) & $\gg50$ \\
\texttt{CFIL+TactileCNN} & 65\% (26/40) & 0.9$\pm$0.5 \\
\texttt{CFIL+Template} & 28\% (11/40) & 1.8$\pm$1.2 \\
\texttt{CFIL} & 30\% (12/40) & 1.7$\pm$0.9 \\
\bottomrule
\end{tabular}
\end{center}
\end{table}

\begin{figure}[t!]
    \centering
    \vspace{-5mm}
    \includegraphics[width=8.4cm]{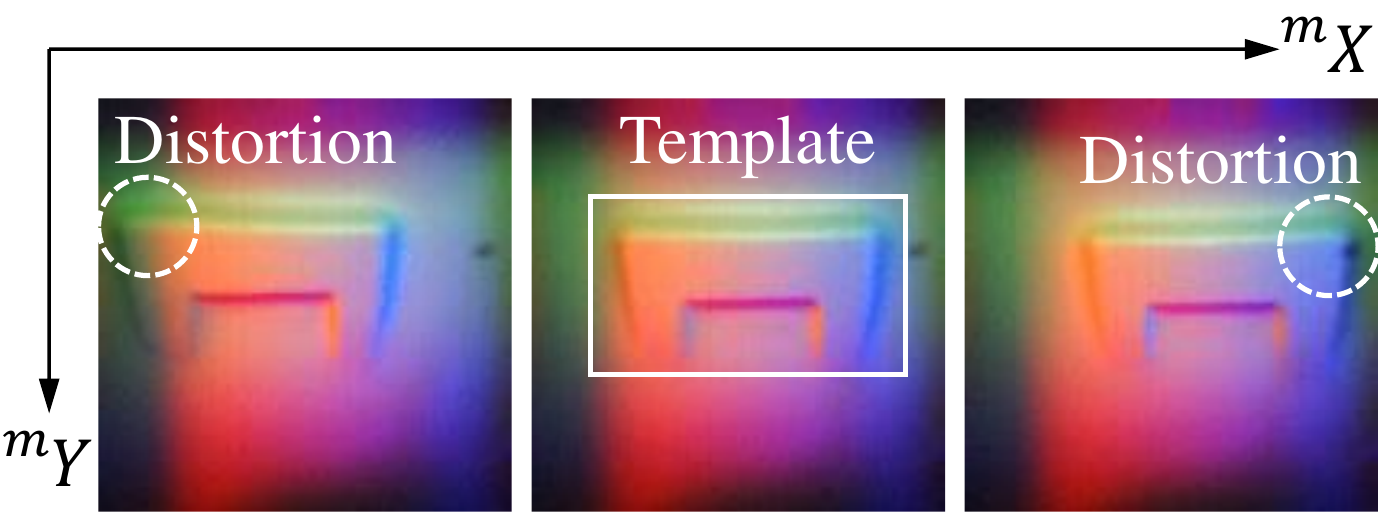}
    \caption{Examples of tactile sensor images.
    Here, $^mX$ and $^mY$ are the axes in an image coordinate system.
    In the template matching experiment, we employed a subset of a tactile sensor image as a template.
    Lighting color and intensity vary depending on the position in the image coordinate system, $^mX$ and $^mY$.
    In the peripheral part of a sensor images, the plug shapes are distorted by lens distortion.}
    \label{fig:tactile_images}
    \vspace{-3mm}
\end{figure}

The proposed method, \texttt{TS-NVAE}, achieved a 100\% success rate and better positioning accuracy ($0.3\pm0.1$ mm), compared to the baselines and ablation studies.

In the NewtonianVAE instances and ablation studies, the connector insertion always failed because the correlations between the physical and latent positions were too low, as discussed in the previous section.

With naive \texttt{CFIL}, the connector insertion succeeded by chance (30\%).
We found that the USB connector insertion failed when the positioning error was greater than $\sim1$ mm.
It is probable that the insertion succeeded when the grasp pose variation was smaller than 1 mm, one third of the maximum grasp pose variation.

We found that \texttt{CFIL+Template} suffered from template matching error, and the success rate was only 28\%.
Fig. \ref{fig:tactile_images} shows some examples of the tactile sensor images.
We used a subset of a tactile sensor image as a template.
In the GelSight-type tactile sensors, the lighting color and intensity vary depending on the position in the image coordinate system, $^mX$ and $^mY$.
In addition, in the peripheral part of the sensor images, the plug shape is distorted by lens distortion.
We found that template matching failed due to these position-dependent image differences.

\texttt{CFIL+TactileCNN} demonstrated a lower success rate (65\%) and lower accuracy ($0.9\pm0.5$ mm) than the proposed method.
This performance degradation can be attributed to three factors.
First, the annotation error for the CNN-based regression model was nonnegligible. 
Second, in the coordinate transformation, $^gT_m$ was measured from the CAD data; however, the mounting error of the gripper and tactile sensor reduced the positioning accuracy.
Note that even small errors can have significant impact on the accuracy of submilllimeter positioning.
Third, the plug pose in the image coordinate system, i.e., $^mT_o$, only considered displacement in the tactile sensor image plane.
Nevertheless, Fig. \ref{fig:tactile_images} shows that the plug was tilted against the image plane.
In the proposed method, such displacement out of the sensor surface was compensated because the insertion position $x_g$ was inferred directly from tactile image latent $z$.

\section{Conclusion}
In this paper, we proposed a general framework for positioning the grasped plug to the insertion position using visual feedback control considering the grasp pose variation without any additional feature engineering, annotation, or calibration.
Furthermore, we induced the domain knowledge on transition uncertainty and grasp pose variation as fixed standard deviation of the Gaussian distribution. 
This domain knowledge is easy to know from robot specification and grasp pose error measurement, and improved the model accuracy. 
Our proposed framework achieved higher performance in the USB connector insertion task in the physical environment, compared to the SOTA CFIL-based goal pose regression with grasp pose compensation using coordinate transformation.

The proposed method has several properties desirable for the industrial application,
such as high accuracy, tunability of the proportional control parameters, sample efficiency, fast and fully offline training, and explainability.

\section{Future Work}
We can extend our method to full-pose proportional control.
Generally, the conventional localization technique like template matching cannot estimate angles accurately.
Therefore, a data-driven method is more important in the full-pose control.

The current control system is disrupted if target objects are framed out, or occluded by other objects,
because the inference models do not use observation histories.
Introducing recurrent architecture into the world models is also an interesting direction.


\addtolength{\textheight}{-2cm}   


\bibliography{iros2022}
\bibliographystyle{ieeetr}

\end{document}